# Adaptive strategy in differential evolution via explicit exploitation and exploration controls

Sheng Xin Zhang[a, b], Wing Shing Chan[b], Kit Sang Tang[b], Shao Yong Zheng[c]

[a]College of Information Science and Technology, Jinan University, Guangzhou, China
[b]Department of Electrical Engineering, City University of Hong Kong, Kowloon, Hong Kong
[c]School of Electronics and Information Technology, Sun Yat-sen University, Guangzhou, China

**Abstract**—Existing multi-strategy adaptive differential evolution (DE) commonly involves trials of multiple strategies and then rewards better-performing ones with more resources. However, the trials of an exploitative or explorative strategy may result in over-exploitation or over-exploration. To improve the performance, this paper proposes a new strategy adaptation method, named explicit adaptation scheme (Ea scheme), which separates multiple strategies and employs them on-demand. It is done by dividing the evolution process into several Selective-candidate with Similarity Selection (SCSS) generations and adaptive generations. In the SCSS generations, the exploitation and exploration needs are learnt by utilizing a balanced strategy. To meet these needs, in adaptive generations, two other strategies, exploitative or explorative is adaptively used. Experimental studies on benchmark functions demonstrate the effectiveness of Ea scheme when compared with its variants and other adaptation methods. Furthermore, performance comparisons with state-of-the-art evolutionary algorithms and swarm intelligence-based algorithms show that EaDE is very competitive.

## 1 Introduction

Exploitation and exploration are two cornerstones of evolutionary algorithms (EAs) [1]. Exploitation refers to the greedy utilization of the currently available information while exploration is the process of discovering new searching areas. As is known, an exploitative way of generating new solutions (i.e. strategy) could efficiently increase accuracy [2]. However, it has high risk of converging to a local minimum. An

Corresponding authors: S. X. Zhang and S. Y. Zheng. (zhangsx@jnu.edu.cn; zhengshaoy@mail.sysu.edu.cn)

explorative strategy can reduce this risk [3, 4], but the accuracy may be unsatisfactory. Multi-strategy methods aim to take advantages of both strategies to improve performance.

In the past decade, multi-mutation strategy [5] based differential evolution (DE) [6-8] has gained much attention from researchers. It is generally believed that introducing multiple alternative strategies in a single algorithm allows adjusting evolution directions as well as evolution scales to meet the exploitation and exploration needs for different searching tasks. Existing multi-strategy techniques can be summarized in the following two categories.

(1) **Adaptive methods.** In this category, adaptive operator selection [9-13] is a popular technique, which involves multiple strategies in the evolution and the past success experience is used for credit assignment to determine the probabilities of the operators that will be used further. This technique has been widely adopted for constructing DE variants, e.g. Strategy adaptation DE (SaDE) [9], Strategy adaptation JADE (SaM-JADE) [10] and Multi-population Ensemble strategy DE (MPEDE) [11] for single-objective optimization. In [12, 13], strategy adaptation has also been extended to multi-objective optimization. A comprehensive survey about strategy ensemble can be found in [5].

(2) **Deterministic methods.** Deterministic methods can also be referred to as non-adaptive methods, which do not utilize feedbacks from the previous search. In the Ensemble of Constraint Handling Techniques (ECHT) [14], different constraint handling methods are employed to generate offspring for its own population. In the Composite DE (CoDE) [15] algorithm, three operators combined with three pairs of control parameters are used to generate three candidates with the fittest one being selected as an offspring. In the Cheap Surrogate Model (CSM) [16], multiple candidates are generated from different operators with the final offspring chosen by a density function. In the Multiple sub-populations Adaptive DE (MPADE) [17], three distinct strategies are assigned to three sub-populations with different fitness values. In the Underestimation-based Multimutation Strategy (UMS) [18], the offspring is determined from multiple candidates by an abstract convex underestimation model. In the Selective-candidate framework with Similarity Selection rule (SCSS) [19] method, each current solution generates multiple candidates using different operations and parameters while the final offspring is determined by the fitness ranking of the current solution and its solution space distance to the candidates.

Although many advances have been achieved, it is still a challenging task to remedy the drawbacks of multi-strategy, i.e. stuck in local minima by an exploitative strategy or over-encouraging exploration by an explorative strategy. In this paper, we propose a new explicit adaptation scheme (i.e. Ea scheme) with the following new features for this task:

(1) Different from existing methods which use multiple strategies at a time, the Ea scheme separates the strategies: one balanced strategy in SCSS generations to optimize, while learning the exploitation and exploration needs; the other two candidates: one exploitative and one explorative are adaptively employed in adaptive generations.

(2) Different from existing methods which adapt strategies based on their online performance compared with those of others, the Ea scheme does not involve the trials of multiple strategies. Instead, it treats the optimization process as a sequence of explicit exploitation and exploration tasks that could be handled by the two candidate strategies, respectively. The prior knowledge of the strategies is pre-studied offline and explicitly used in the scheme.

The contributions of this paper are multi-fold: (1) we construct and study the exploitation and exploration capabilities of the strategies; (2) we propose the innovative Ea scheme and demonstrate its advantages by comparison with other adaptation methods; and (3) we construct the EaDE algorithm with state-of-the-art performance.

The rest of the paper is organized as follows: Section 2 reviews related works and discusses the novelty of the Ea scheme. Section 3 describes the proposed method in detail. Section 4 presents the experimental validations together with discussions while Section 5 concludes this work.

## 2 Backgrounds
### 2.1 Strategy adaptation in DE

In this paper, strategy refers to the way of generating new solutions from the parent solutions, including mutation and crossover operations of DE. While in DE literature [7, 8], mutation strategy, especially mutation strategy adaptation is more widely studied. In DE with Global and Local mutation (DEGL) [20], global and local mutation strategies are combined to balance exploitation and exploration using an adaptive weighting factor. In SaDE [9], four mutation strategies are adaptively used based on their past success and fail experiences. In SaM-JADE [10], four strategies are indexed, and the indices are regarded as parameters for adaptation. In Adaptive strategy (Adap_SS) [21], probability matching and adaptive pursuit techniques are used to calculate the probabilities of strategies. In Ensemble of Parameters and mutation Strategies DE (EPSDE) [22], parameters and strategies are distributed to solutions based on their successful and fail experiences. In Zoning Evolution of Parameters based DE (ZEPDE) [23], mutation strategies and parameters are adjusted based on their fitness improvements. In MPEDE [11], three mutation strategies are assigned to three small sub-populations and the best-performing one is rewarded a large sub-population. In MOEA/D-FRRMAB [12], four mutation strategies compete based on multi-armed bandits. In Multistage DE (UMDE) [24], a strategy pool is constructed based on the evolution stage and at the same stage, different strategies compete based on their fitness improvements. In multiple variants coordination DE (MVCDE) [25], multiple DE variants compete based on their contributions. In Chaotic local search-based DE (CDE) [26], different chaotic maps are adaptively used to generate new solutions based on their success rates. In the above methods [9-12, 21-26], successful strategies/methods generally occupy more computation resources. In Multi-topology DE (MTDE) [27], new solutions are generated using different topologies determined by different fitness values. In Neighborhood-based DE (NDE) [28], two mutation strategies are adaptively associated with superior and inferior solutions for exploitation and exploration purposes, respectively. In Multi-layer Competitive-Cooperative DE (MLCCDE) [29], superior solutions use multiple methods to generate solutions while inferior solutions adaptively choose one method, which is designed with the consideration of exploitation and exploration trade-off.

**2. 2 Novelty of the Ea scheme**

Differences of the proposed Ea scheme compared to the previous adaptation methods are illustrated in Fig. 1: (1) Previous methods usually involve multiple strategies with each strategy assigned to a portion of the population. In contrast, the Ea scheme employs one strategy at a time for the whole population; (2) Previous methods usually trial multiple strategies and reward the better-performing strategy. In determining whether a strategy is better-performing or not, the metric is commonly greedy. For example, in Fig. 1(a), three strategies compete at time $t$ and assume that the green one performs better, then at time $t+1$, it will be rewarded with more resources. Differently, Ea scheme uses a balanced strategy to explicitly detect exploitation/exploration needs (Fig. 1(b)). Afterwards, the entire population switches to an exploitative/explorative strategy. A performance comparison of these methods will be presented in Section 4.3.

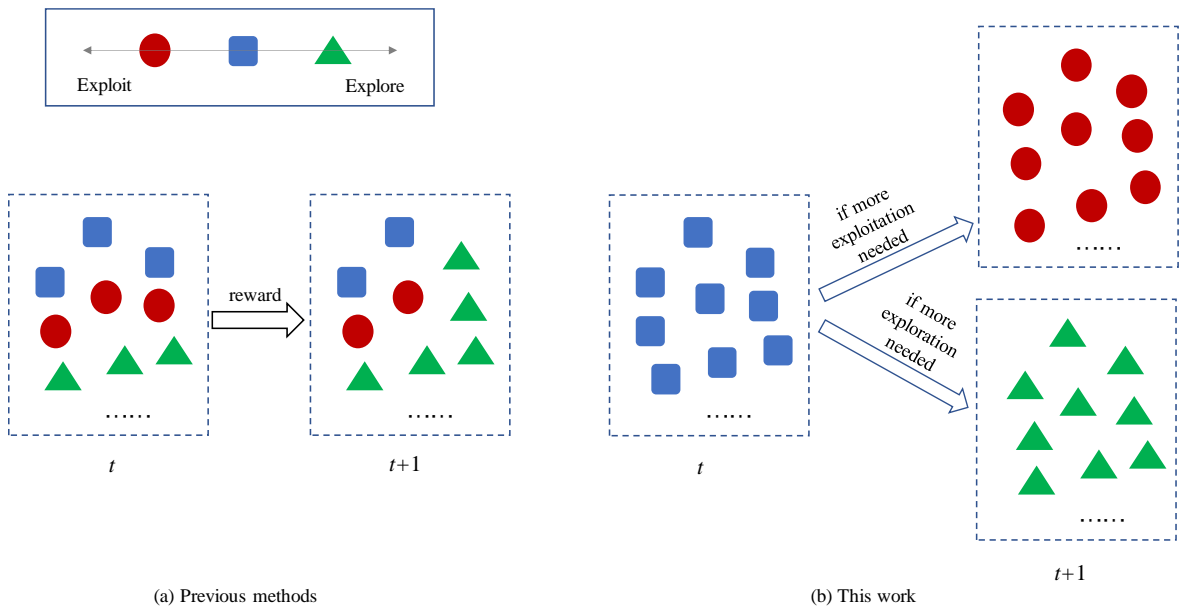

Fig. 1 Illustration of the proposed and the previous adaptation methods. A blue square, red circle and green triangle denote that a balanced, exploitative and explorative strategy is assigned to a solution respectively.

## 2.3 Detection of exploitation/exploration need: SCSS method

In Selective-candidate with Similarity Selection rule based L-SHADE (SCSS-L-SHADE) [19], two independent reproductions are performed to generate two candidates $\vec{u}_{i,G}^{\,m}$ $\{m = 1, 2\}$ for each current solution $\vec{x}_{i,G}$ at each generation $G$ and the offspring is determined by the following similarity selection rule (**Algorithm 1**):

---

**Algorithm 1**: Similarity selection rule

---

    **If** $rand_i(0,1) \times 2 \times GD > rank(i)/NP$

        Select the closest candidate from $\vec{u}_{i,G}^{\,m}$ $\{m = 1, 2\}$ for current solution $\vec{x}_{i,G}$;

    **Else**

        Select the farthest candidate from $\vec{u}_{i,G}^{\,m}$ $\{m = 1, 2\}$ for current solution $\vec{x}_{i,G}$;

    **End If**

---

Here, $rand_i(0,1)$ is a uniformly distributed random number within (0,1) for each individual $i$, $rank(i)$ is the fitness ranking (the smaller, the better quality) and $NP$ is the current population size. In this rule, the superior solution prefers closer candidate, while the inferior solution prefers the farther one. $GD$ (greedy degree) value controls the greediness of the selection rule. The larger $GD$, the more current solutions select closer candidates and consequently the algorithm becomes more exploitative. The average distance of offspring from parent solutions in SCSS-L-SHADE with $GD = 0.5$ (denoted as SCSS-L-SHADE_GD0.5) and original L-SHADE is shown in Fig. 2.

It was reported in [19] that SCSS-L-SHADE_GD0.5 performs better than L-SHADE on a wide variety of benchmarks and real-world problems for its synthesis of exploitation (superior solutions) and exploration (inferior solutions). Interestingly, the motivation behind strategy adaptation is also to detect and meet the exploitation and exploration needs (EEN). This motivates us to utilize it as a tool for EEN detection.

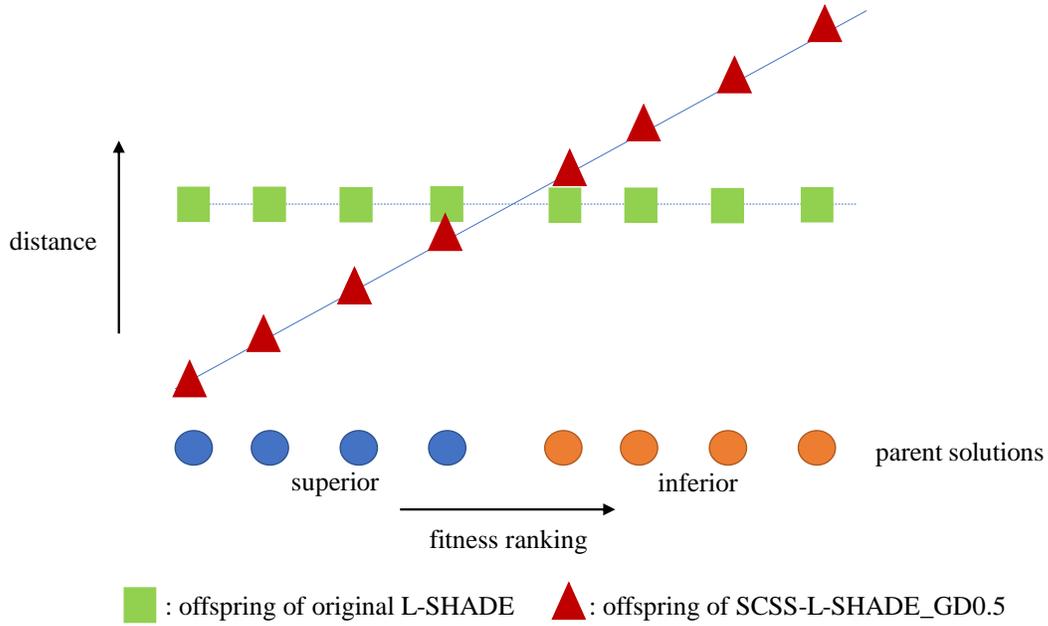

Fig. 2 Average distance of offspring from parent solutions in SCSS-L-SHADE_GD0.5 and the original L-SHADE (A population with 8 solutions is shown as an example).

## 3 EaDE (Explicitly adaptive DE)

### 3.1 Exploitation and exploration capabilities of the strategies in EaDE

Besides SCSS-L-SHADE_GD0.5 (marked as S1), two other strategies are maintained in EaDE, i.e. SCSS-L-SHADE with $GD = 0.1$ and SCSS-L-CIPDE with $GD = 0.9$, denoted as SCSS-L-SHADE_GD0.1 (marked as S2) and SCSS-L-CIPDE_GD0.9 (marked as S3) respectively. The difference between SCSS-L-SHADE and SCSS-L-CIPDE lies in the mutation and crossover operations. The former employs the "current-to-*p*best/1" mutation and classic binomial crossover [30] while the latter uses the collective information powered (CIP) mutation and hybrid crossover [31]. Detailed descriptions of these operations can be found in the supplemental file. The principle behind employing these two strategies is that SCSS-L-CIPDE_GD0.9 is relatively exploitative while SCSS-L-SHADE_GD0.1 is relatively explorative. To demonstrate these properties, we employed the three strategies to independently sample three offspring populations $\vec{u}_{i,G}$ on the same parent population at each generation $G$ on twenty-eight 30-dimensional (30-D) CEC2013 benchmark functions [32].

Define diversity of the sampled populations as

$$Div_G = \frac{1}{NP}\sum_{i=1}^{NP}\|\vec{u}_{i,G} - \bar{u}_G\| \tag{1}$$

where $\|\vec{u}_{i,G} - \bar{u}_G\|$ is the Euclidian distance between $\vec{u}_{i,G}$ and $\bar{u}_G$, and $\bar{u}_G = \frac{1}{NP}\sum_{i=1}^{NP}\vec{u}_{i,G}$.

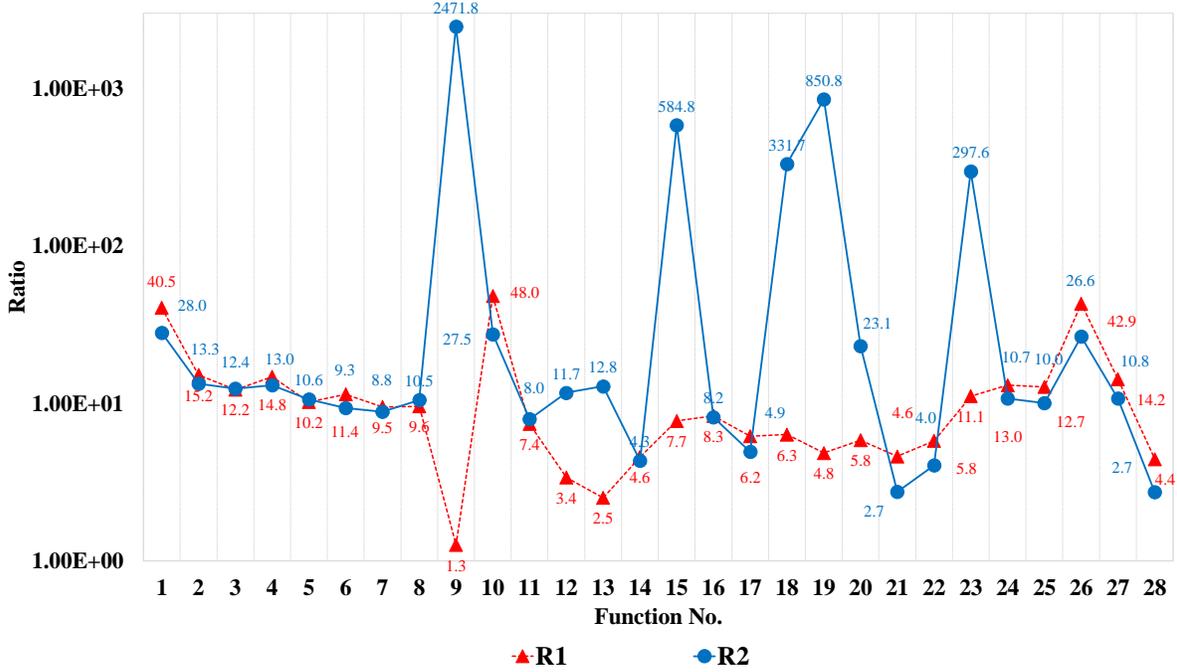

Fig. 3 $R1$ and $R2$ values on twenty-eight 30-D CEC2013 functions with maximum function evaluations 10000×D. For each function, 30 trials were performed.

Table 1 Performance ranking of the $GD$ values for SCSS-L-CIPDE (on *Set* 1) and SCSS-L-SHADE (on *Set* 2). The smaller ranking value, the better.

| GD | 0.0 | 0.1 | 0.2 | 0.3 | 0.4 | 0.5 | 0.6 | 0.7 | 0.8 | 0.9 | 1.0 |
|---|---|---|---|---|---|---|---|---|---|---|---|
| *Set*1 | 10.50 | 10.00 | 8.90 | 7.20 | 6.20 | 3.80 | 3.80 | 3.80 | 2.60 | **2.20** | 7.00 |
| *Set*2 | 5.83 | **4.12** | 4.20 | 4.29 | 4.41 | 4.91 | 4.75 | 6.33 | 8.00 | 8.91 | 10.20 |

In this experiment, we intended to compare the exploitation and exploration capabilities of S1 with S2 and S1 with S3, respectively. To this end, we compared the diversity of the population generated by S1 and S2, S1 and S3 respectively at each generation and then counted accumulatively. The average times that the population generated by S1 has smaller/larger diversity than population by S2 are denoted as $T_{S1<S2}$ and $T_{S1>S2}$ respectively. Thus, the ratio $R1 = T_{S1<S2}/T_{S1>S2}$ represents the relative greediness of the two strategies. If the value is larger than 1, it means S1 is more exploitative than S2. Similarly, we used ratio $R2 = T_{S3<S1}/T_{S3>S1}$ for the comparison of S1 and S3.

The obtained $R1$ and $R2$ values on the total 28 functions are shown in Fig. 3. It is seen that $R1$ and $R2$ are consistently larger than 1 on all the functions, indicating that S1 is more exploitative than S2 and S3 is more exploitative than S1 respectively. We have the following exploitation capability ranking: SCSS-L-CIPDE_GD0.9 > SCSS-L-SHADE_GD0.5 > SCSS-L-SHADE_GD0.1 and thus the exploration capability ranking: SCSS-L-CIPDE_GD0.9 < SCSS-L-SHADE_GD0.5 < SCSS-L-SHADE_GD0.1, where ">" means stronger than and "<" means weaker than.

With respect to the choice of *GD* values, firstly, we classified the 30-D CEC2013 functions into two sets, i.e. *Set 1* (larger *GD* performs better) and *Set 2* (smaller *GD* performs better). Thus, these two sets can be used to examine the exploitation and exploration capabilities of an algorithm respectively. Then we test SCSS-L-CIPDE and SCSS-L-SHADE with eleven *GD* values (from 0.0 to 1.0 with step of 0.1) on *Set 1* and *Set 2*, respectively. The overall performance ranking by Friedman's test [33] is shown in Table 1. As suggested by the results, *GD* = 0.9 and 0.1 are respectively chosen for SCSS-L-CIPDE and SCSS-L-SHADE.

### 3.2 Explicit adaptation (Ea scheme)

The division of different generations and the characteristics of the associated strategies in the proposed Ea scheme are illustrated in Fig. 4. The entire evolution is segmented into several non-overlapped intervals with equal number of generations. In each interval, there are SCSS generations and adaptive generations with sizes of *LEN* and *K*×*LEN* respectively, where *K* is an integer.

In SCSS generations, SCSS-L-SHADE_GD0.5 is performed and the total fitness improvements of superior and inferior parts are calculated respectively as:

$$IMP\_S = \sum_{g=1}^{LEN} \sum_{rank(i)=1}^{\lfloor NP/2 \rfloor} \Delta f_i \quad (2)$$

$$IMP\_I = \sum_{g=1}^{LEN} \sum_{rank(i)=\lfloor NP/2 \rfloor+1}^{NP} \Delta f_i \quad (3)$$

Where $\Delta f_i = \begin{cases} f(\vec{x}_{i,G}) - f(\vec{u}_{i,G}) & \text{if } f(\vec{u}_{i,G}) < f(\vec{x}_{i,G}) \\ 0 & \text{otherwise} \end{cases}$, $\lfloor \cdot \rfloor$ represents a floor function and $\lceil \cdot \rceil$ is a ceiling function.

In the adjacent adaptive generations, **Algorithm 2** is performed. The principle behind this design is as follows: According to Section 2.3, superior and inferior solutions in SCSS-L-SHADE_GD0.5 are responsible for exploitation and exploration tasks respectively. A larger contribution (i.e. total fitness improvements) of the superior part compared to that of the inferior part indicates that the current stage may need more exploitation attempts. Therefore, an exploitative strategy SCSS-L-CIPDE_GD0.9 will be used. Otherwise, if the inferior part contributes more, an explorative strategy may be more suitable. Effectiveness of the proposed Ea scheme will be verified in Section 4.1.

Regarding the time complexity, it can be seen that it is relatively low for the proposed method since it only involves $O(NP)$ summation calculations at each generation.

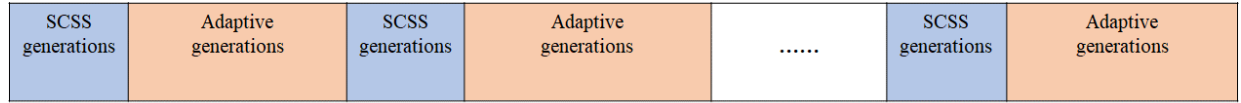

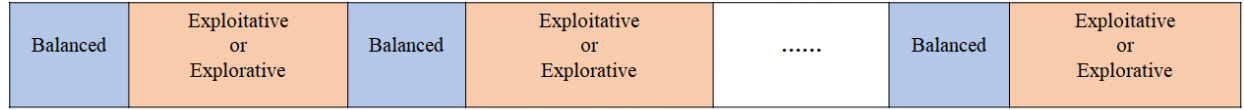

Fig. 4 Illustration of the Ea scheme, (a): division of generations; (b): characteristics of the associated strategies.

---

**Algorithm 2**: **Ea scheme**

---

    **If** $IMP\_S > IMP\_I$

        Run SCSS-L-CIPDE_GD0.9

    **Elseif** $IMP\_S < IMP\_I$

        Run SCSS-L-SHADE_GD0.1

    **Elseif** $IMP\_S = IMP\_I$

        Run a random strategy of the above two.

    **End If**

---

### 3.3 Detection of evolution difficulty

With "current-to-$p$best/1" mutation, in the early evolution stage, compared to superior solutions, the inferior solutions can be easily guided by the top-$p$ fittest solutions and easier to be improved. Thus, at this stage, it may be unfair to compare these two parts. To address this problem, we propose a detection mechanism (DM) to detect evolution difficulty and combine it with the Ea scheme. At the beginning, Ea is not triggered, i.e. Trigger = 0. Then the total fitness improvements of superior and inferior parts within every $Q$ generations ($Q$ is set to 10 in this paper) are recorded respectively, denoted as $FI\_S$ and $FI\_I$. At the early stage, $FI\_I$ tends to be larger than $FI\_S$. Once $FI\_S$ is larger than $FI\_I$, it means that the current evolution stage becomes difficult. The Ea scheme is then triggered, i.e. Trigger = 1 and adopted until the end of the search. Otherwise, SCSS-L-SHADE_GD0.5 is employed and Trigger = 0.

To demonstrate the effectiveness of DM, four classic benchmark functions, including Sphere, Sum of different powers, Schwefel and Rastrigin functions are used. Their mathematical expressions and plots with two variables are shown in Table 2 and Fig. 5 respectively. As is known, Schwefel and Rastrigin functions are more difficult to solve than Sphere and Sum of different powers functions since they have many local minima.

Table 3 reports the trigger time on these four functions, where $t$ is the run-time of a single trial. It is seen that on two relatively simple functions, i.e. Sphere and Sum of different powers functions, Ea triggers at a very late stage, i.e. $0.92t$ and $0.94t$ respectively. While on Schwefel and Rastrigin functions, Ea triggers at $0.32t$ and $0.19t$ respectively. Considering the function difficulty, this observation confirms that DM could effectively detect the evolution difficulty.

Table 3 also collects the performance with and without DM. It is seen that the algorithm with DM performs significantly better on two simple functions. The reason is that DM could prevent over-exploration in the early stage.

Regarding the setting of $Q$, Table 4 shows the comparison results of the standard setting $Q = 10$ with others. It is seen that $Q = 10$ is the best choice. It performs better than smaller $Q$ settings mainly on the simple sphere and sum of different powers functions while larger $Q$ settings on the relatively difficult Rastrigin function.

Table 2 Mathematical expressions of the functions

| Function | Definition | Search range |
|---|---|---|
| Sphere Function | $f_1(x) = \sum_{i=1}^{D} x_i^2$ | $[-100, 100]^D$ |
| Sum of different powers Function | $f_2(x) = \sum_{i=1}^{D} \lvert x_i \rvert^{i+1}$ | $[-1, 1]^D$ |
| Schwefel Function | $f_3(x) = 418.9829 \times D - \sum_{i=1}^{D} x_i \sin(\sqrt{\lvert x_i \rvert})$ | $[-500, 500]^D$ |
| Rastrigin Function | $f_4(x) = \sum_{i=1}^{D} [x_i^2 - 10\cos(2\pi x_i) + 10]$ | $[-5, 5]^D$ |

Table 3 Trigger time on 30-D functions and performance contribution of DM. Maximum function evaluations = 10000×D. For each function, 30 trials were performed. Wilcoxon signed-rank test with 5% significance level is used to determine the significance.

| Function | Trigger at | Performance Without DM | Performance With DM | Significance. |
|---|---|---|---|---|
| Sphere Function | $0.92t$ | 9.86E-55±2.91E-54 | **2.42E-78±1.66E-77** | better |
| Sum of different powers Function | $0.94t$ | 1.96E-123±1.16E-122 | **3.69E-154±2.19E-153** | better |
| Schwefel Function | $0.32t$ | 3.82E-04±3.57E-13 | 3.82E-04±0.00E+00 | similar |
| Rastrigin Function | $0.19t$ | 1.15E-15±3.19E-15 | 6.27E-16±1.58E-15 | similar |

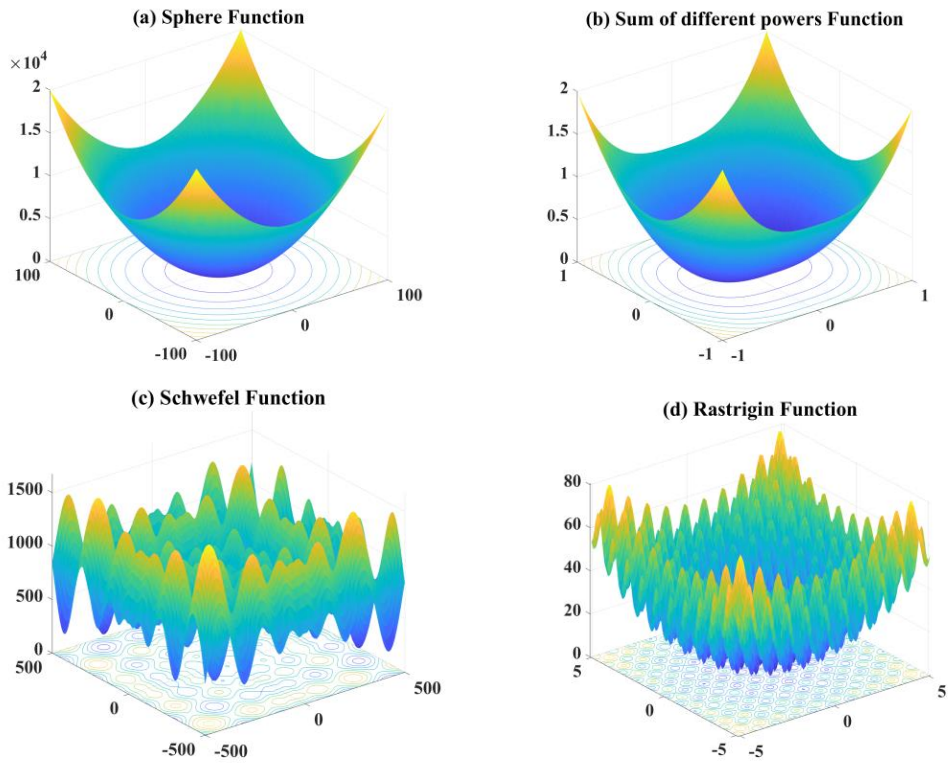

Fig. 5 Plots of 2-D functions: (a) Sphere function; (b) Sum of different powers function; (c) Schwefel function and (d) Rastrigin function. Two horizontal axes are two variables while the vertical axis is the fitness.

Table 4 Comparison results of $Q = 10$ with other settings

| $Q$ | 1 | 5 | 20 | 40 | 80 | 160 | 320 | 640 | 1280 | 2560 |
|---|---|---|---|---|---|---|---|---|---|---|
| Sphere Function | W | W | T | T | T | T | T | T | T | T |
| Sum of different powers Function | W | T | T | T | T | T | T | T | T | T |
| Schwefel Function | T | T | T | T | T | T | T | T | T | T |
| Rastrigin Function | T | W | W | W | W | W | W | W | W | W |

W: win, T: tie, L: lose

### 3.4 The complete EaDE algorithm

Combining the above methods, the complete EaDE algorithm is described in **Algorithm 3** and illustrated in Fig. 6. At the beginning, SCSS-L-SHADE_GD0.5 is run and the Ea scheme is not triggered (lines 1 and 2 in Algorithm 3). Line 3 detects the evolution difficulty and if it enters a relatively difficult stage, Ea scheme will be triggered (lines 6-13).

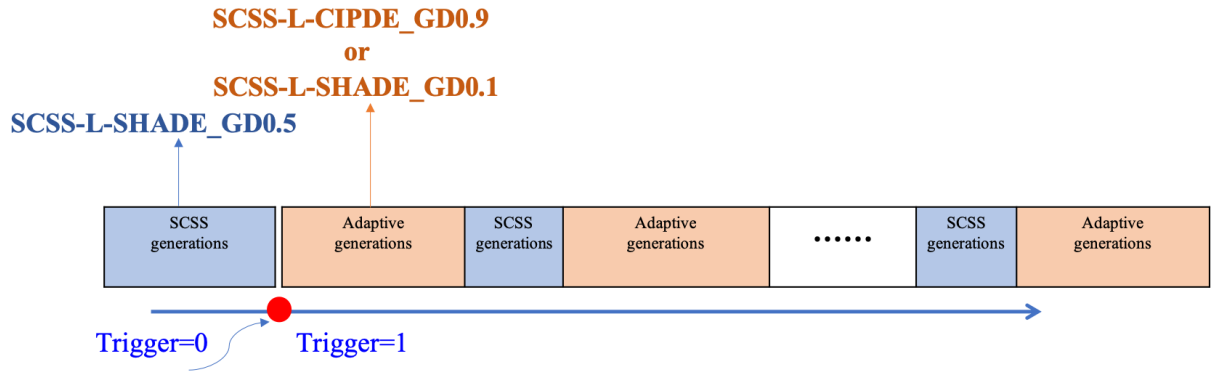

Fig. 6 Illustration of EaDE.

---

**Algorithm 3**: **EaDE**

---

1: Trigger = 0;

2: Run SCSS-L-SHADE_GD0.5;

3: Respectively record the total fitness improvements of superior and inferior solutions within every $Q$ generations, denoted as *FI_S* and *FI_I*.

4: **If** *FI_S* > *FI_I*

5:  Trigger = 1,

--------------------------------**Ea scheme**-----------------------------------------

6:  For each interval, calculate *IMP_S* and *IMP_I*;

7:  **If** *IMP_S* > *IMP_I*

8:   Run SCSS-L-CIPDE_GD0.9

9:  **Elseif** *IMP_S* < *IMP_I*

10:   Run SCSS-L-SHADE_GD0.1

11: **Elseif** *IMP_S* = *IMP_I*

12:   Run a random strategy of the above two.

13: **End If**

-------------------------------------------------------------------------------------

14: **Else**

15:  Trigger = 0, run SCSS-L-SHADE_GD0.5;

16: **End If**

---

## 4 Simulation

In this section, we conduct experiments to verify the effectiveness of the proposed method. The structure is organized as follows: in Sections 4.1-4.3, the Ea scheme is compared respectively with three variants, three components and two other adaptation methods to verify its effectiveness. In Sections 4.4 and 4.5, EaDE is compared with state-of-the art optimization algorithms to demonstrate its performance. Finally, in Section 4.6, we investigate the performance sensitivity of EaDE to parameters *LEN* and *K* and present some discussions on the number of strategies, respectively.

The CEC2013 test suite [32] with twenty-eight minimization benchmark functions is considered, as shown in Table 5. Performance of algorithms are measured by the final obtain best solution. Following [32], solutions smaller than $10^{-8}$ are reported as 0. For each function, 51 trials are performed, each with $10000 \times D$ maximum function evaluations (FES). The 5% significance level Wilcoxon signed-rank test [33] is used to compare the performance. When the compared algorithm is significantly worse than, similar to or better than the algorithm under consideration, we mark it as "−", "=" and "+", respectively. Parameters settings for EaDE: The initial population size is set to $18 \times D$ and linearly decreased to 4 at the end, this setting is kept the same as SCSS-L-SHADE [19]. The size of SCSS generations *LEN* is set to 30 and the size of adaptive generations is set to $K \times LEN = 2 \times 30 = 60$.

Table 5 The CEC2013 test suite

| F1 | Sphere Function | F11 | Rastrigin's Function |
|---|---|---|---|
| F2 | Rotated High Conditioned Elliptic Function | F12 | Rotated Rastrigin's Function |
| F3 | Rotated Bent Cigar Function | F13 | Non-Continuous Rotated Rastrigin's Function |
| F4 | Rotated Discus Function | F14 | Schwefel's Function |
| F5 | Different Powers Function | F15 | Rotated Schwefel's Function |
| F6 | Rotated Rosenbrock's Function | F16 | Rotated Katsuura Function |
| F7 | Rotated Schaffers F7 Function | F17 | Lunacek Bi_Rastrigin Function |
| F8 | Rotated Ackley's Function | F18 | Rotated Lunacek Bi_Rastrigin Function |
| F9 | Rotated Weierstrass Function | F19 | Expanded Griewank's plus Rosenbrock's Function |
| F10 | Rotated Griewank's Function | F20 | Expanded Scaffer's F6 Function |
| F21-F28 | Composition Function | | |

### 4.1 Effectiveness of the Ea scheme: Comparison with three variants

Firstly, we construct the following three variants to verify the effectiveness of the Ea scheme.

*Variant-oppo*: It is an opposite version of Ea, as follows:

    **If** *IMP_S > IMP_I*

        Run SCSS-L-SHADE_GD0.1

    **Elseif** *IMP_S < IMP_I*

        Run SCSS-L-CIPDE_GD0.9

    **Elseif** *IMP_S = IMP_I*

        Run a random strategy of the above two.

    **End If**

*Variant-random*: Different from Ea, in adaptive generations, SCSS-L-SHADE_GD0.1 and SCSS-L-CIPDE_GD0.9 are randomly used.

*Variant-TAE*: It is an *adaptive with trial-and-error* variant, as illustrated in Fig. 7 and described as follows:

In trial generations, SCSS-L-CIPDE_GD0.9, SCSS-L-SHADE_GD0.5 and SCSS-L-SHADE_GD0.1 have an equal chance to be used. Then record the total fitness improvements contributed by them respectively, denoted as *IMP_CIP* and *IMP_SHA_0.5* and *IMP_SHA_0.1*.

In the adjacent adaptive generations,

    **If** *IMP_CIP* is the unique largest in {*IMP_CIP*, *IMP_SHA_0.5*, *IMP_SHA_0.1*}

        Run SCSS-L-CIPDE_GD0.9

    **Elseif** *IMP_SHA_0.1* is the unique largest in {*IMP_CIP*, *IMP_SHA_0.5*, *IMP_SHA_0.1*}

        Run SCSS-L-SHADE_GD0.1

    **Elseif** *IMP_SHA_0.5* is the unique largest in {*IMP_CIP*, *IMP_SHA_0.5*, *IMP_SHA_0.1*}

        Run SCSS-L-SHADE_GD0.5

    **Elseif** more than one largest value in {*IMP_CIP*, *IMP_SHA_0.5*, *IMP_SHA_0.1*}

        Run a random strategy with the largest value.

    **End If**

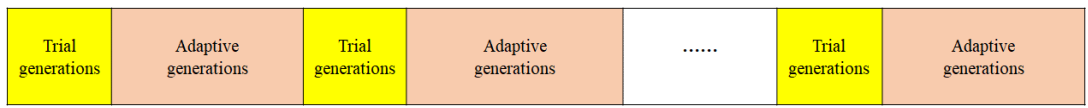

Fig. 7 Illustration of Variant-TAE.

For a direct comparison, except the above differences, other settings are kept the same as EaDE. The pseudo-code is presented in the supplemental file. Table S1 in the supplemental file reports the mean (standard deviations) and comparison results for 10-D, 30-D, 50-D and 100-D cases. The results are summarized in Table 6 and discussed as follows:

(1) EaDE performs significantly better than Variant-oppo, winning in 41 (=6+11+13+11) cases and losing in 1 case. This result confirms the effectiveness of Ea scheme.

(2) EaDE also outperforms Variant-random with the "win/tie/lose" results of "2/26/0", "6/22/0", "7/21/0" and "7/21/0" in 10-D, 30-D, 50-D and 100-D cases respectively with no function losing to Variant-random. Meanwhile, considering the results of Variant-oppo, it is seen that Variant-random performs better than Variant-oppo, further indicating that Variant-oppo is not an appropriate choice. In principle, Variant-random has 50% probability in using the same strategy in each adaptive generations as EaDE, while Variant-oppo uses exactly the opposite strategy. This explains the difference in performance: Variant-oppo is worse than Variant-random and Variant-random is worse than EaDE.

(3) EaDE without trial-and-error is significantly better than Variant-TAE with trial-and-error in 42 (=2+13+13+14) cases and worse in 10 (=2+2+2+4) cases. In Section 1, we claimed that "*it is still a challenge task to remedy the drawbacks, i.e. stuck in local minima by an exploitative strategy or over-encouraging exploration by an explorative strategy.*". When introducing a new strategy, it could bring advantages when solving some problems, however, it may also introduce weaknesses in other kinds of problems. To demonstrate this, we consider two 30-D functions, F13 and F22. Table 7 collects the results of the compared algorithms on these functions. As seen from Table 7, among the three strategies, SCSS-L-CIPDE_GD0.9 and SCSS-L-SHADE_GD0.1 have an advantage on only one of the two functions. Variant-TAE employs the three strategies to try and determine the best one to use in adaptive generations. For F13, it performs better than all the strategies, meaning that Variant-TAE could adapt to appropriate strategies on this function. While for EaDE, it not only performs better than the baselines but also outperforms Variant-TAE. For F22, Variant-TAE could not eliminate the disadvantage of involving SCSS-L-CIPDE_GD0.9 and as a result, it loses to SCSS-L-SHADE_GD0.1 although it performs better than the other two strategies. While for EaDE, it overcomes the drawback of SCSS-L-CIPDE_GD0.9. This can be explained by the fact that EaDE employs SCSS-L-CIPDE_GD0.9 only when it is needed while Variant-TAE includes it in the trial-and-error process.

Table 6 Comparison results of EaDE with the variants
on 10-D, 30-D, 50-D and 100-D CEC2013 benchmark set

| | EaDE vs. | Variant-oppo | Variant-random | Variant-TAE |
|---|---|---|---|---|
| | win | 6 | 2 | 2 |
| 10-D | tie | 21 | 26 | 24 |
| | lose | 1 | 0 | 2 |
| | win | 11 | 6 | 13 |
| 30-D | tie | 17 | 22 | 13 |
| | lose | 0 | 0 | 2 |
| | win | 13 | 7 | 13 |
| 50-D | tie | 15 | 21 | 13 |
| | lose | 0 | 0 | 2 |
| | win | 11 | 7 | 14 |
| 100-D | tie | 17 | 21 | 10 |
| | lose | 0 | 0 | 4 |

Table 7 Experiment results on 30-D F13 and F22

| | SCSS-L-CIPDE_GD0.9 | SCSS-L-SHADE_GD0.1 | SCSS-L-SHADE_GD0.5 | Variant-TAE | EaDE |
|---|---|---|---|---|---|
| F13 | 6.04 | 8.53 | 6.89 | 2.31 | 1.63 |
| | (6.30) | (3.64) | (3.71) | (2.67) | (2.40) |
| F22 | 115.61 | 106.85 | 108.10 | 107.14 | 106.86 |
| | (2.97) | (1.84) | (1.50) | (1.35) | (1.73) |
| vs.Variant-TAE (F13) | − | − | − | | |
| vs. EaDE (F13) | − | − | − | − | |
| vs.Variant-TAE (F22) | − | + | − | | |
| vs. EaDE (F22) | − | = | − | − | |

## 4.2 Effectiveness of the Ea scheme: Comparison with three components

Further, EaDE is compared with three components, i.e. SCSS-L-CIPDE_GD0.9, SCSS-L-SHADE_GD0.1 and SCSS-L-SHADE_GD0.5 on 10-D, 30-D, 50-D and 100-D functions. The experimental results are presented in Table S2 in the supplemental file and summarized in Table 8.

As seen from Table 8, compared with SCSS-L-SHADE_GD0.5, EaDE is significantly better on most of the functions, winning in 42 (=8+12+12+10) and losing in 4 (=0+0+1+3) cases. This indicates that the proposed Ea scheme is effective in adjusting the exploitation and exploration capabilities of SCSS-L-SHADE_GD0.5, leading to a much better performance. Compared with the other two strategies, EaDE also performs better. Specifically, SCSS-L-CIPDE_GD0.9 outperforms EaDE on F18 in both 30-D and 50-D cases. In most other cases, it is worse than EaDE. SCSS-L-SHADE_GD0.1 performs better than EaDE in 50-D F16 while on the rest functions, it is worse than or similar to EaDE.

Table 8 Comparison results of EaDE with the components on 10-D, 30-D, 50-D and 100-D CEC2013 benchmark set

|  | EaDE vs. | SCSS-L-CIPDE_GD0.9 | SCSS-L-SHADE_GD0.1 | SCSS-L-SHADE_GD0.5 |
|---|---|---|---|---|
| 10-D | win | 5 | 7 | 8 |
|  | tie | 19 | 21 | 20 |
|  | lose | 4 | 0 | 0 |
| 30-D | win | 18 | 10 | 12 |
|  | tie | 9 | 18 | 16 |
|  | lose | 1 | 0 | 0 |
| 50-D | win | 18 | 9 | 12 |
|  | tie | 5 | 18 | 15 |
|  | lose | 5 | 1 | 1 |
| 100-D | win | 18 | 10 | 10 |
|  | tie | 7 | 16 | 15 |
|  | lose | 3 | 2 | 3 |

To investigate the usages of strategies, the trajectory of the employed strategies at different adaptive generations intervals on 30-D F13 and F22 is shown in Fig. 8. As seen from Table 7, SCSS-L-CIPDE_GD0.9 is better than SCSS-L-SHADE_GD0.1 when solving F13. Interestingly, in EaDE, SCSS-L-CIPDE_GD0.9 is not always used for this function and consequently, EaDE could outperform SCSS-L-CIPDE_GD0.9. For F22, Table 7 shows that SCSS-L-SHADE_GD0.1 is a better choice compared to SCSS-L-CIPDE_GD0.9. In EaDE, SCSS-L-SHADE_GD0.1 is selected in most of the adaptive generations. This may explain the result that EaDE is competitive to SCSS-L-SHADE_GD0.1 while outperforming SCSS-L-CIPDE_GD0.9.

In conclusion, EaDE could effectively adapt to appropriate strategies. It not only outperforms SCSS-L-SHADE_GD0.5 but also the other two strategies in most of the cases.

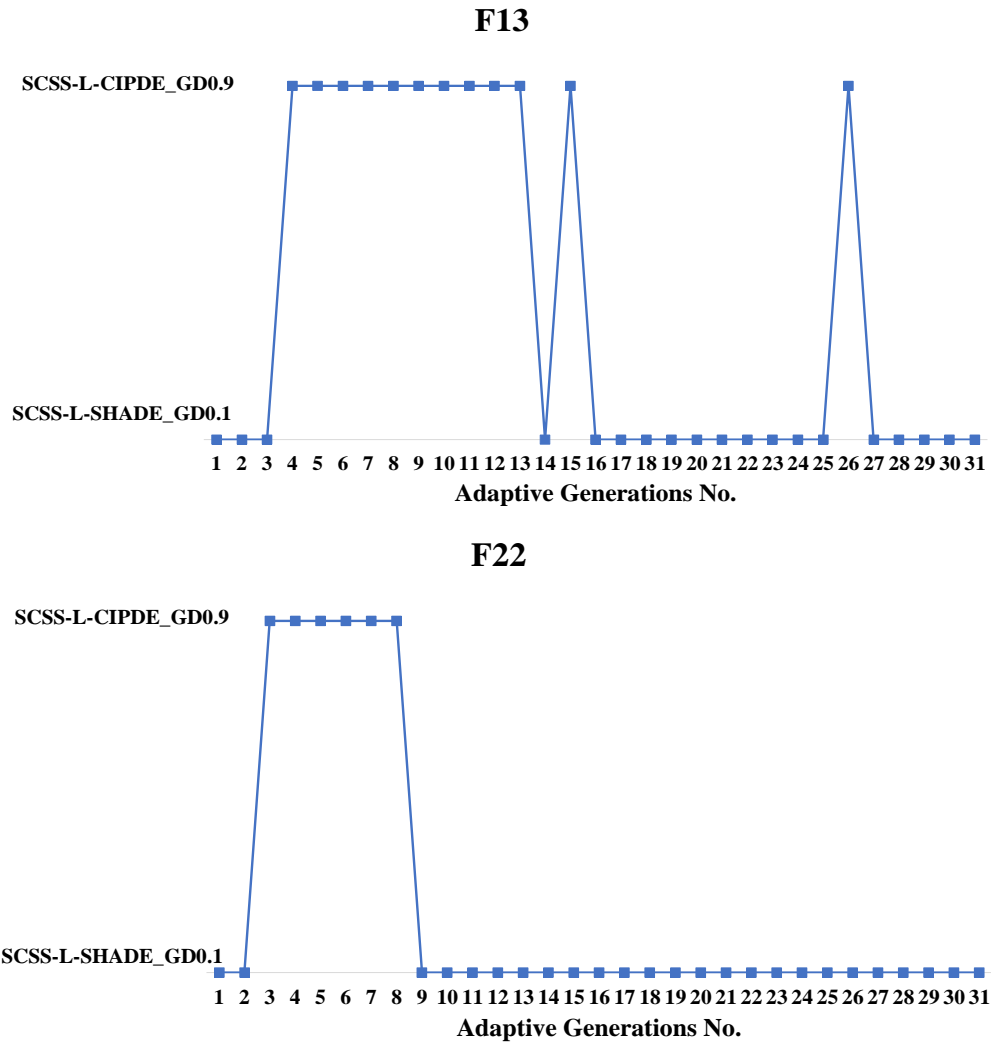

Fig. 8 Usages of strategies at different adaptive generations on 30-D F13 and F22.

### 4.3 Effectiveness of the Ea scheme: Comparison with other adaptation methods

As illustrated in Section 2.2 and Fig. 1, differences between the Ea scheme and previous methods can be clearly observed. It would also be interesting to compare their performance. To this end, adaptation methods proposed in SaDE [9] and SaM-JADE [10] are respectively applied to the three strategies, i.e. SCSS-L-CIPDE_GD0.9, SCSS-L-SHADE_GD0.1 and SCSS-L-SHADE_GD0.5. The resultant algorithms, named Variant-Sa and Variant-SaM are compared with EaDE.

Experimental results are presented in Table S3 in the supplemental file. From the results summarized in Table 9, it is seen that EaDE performs better than both Variant-Sa and Variant-SaM. For instance, with respect to the 30-D case, EaDE outperforms Variant-Sa and Variant-SaM in 6 (F3, F8, F14, F16, F17, F22), 16 cases (F2, F3, F6, F7, F10, F11, F13, F14, F16, F17, F22, F24-F28) and underperformed in 1 (F20), 1 case (F9) respectively. In the 50-D case, compared with Variant-Sa and Variant-SaM, EaDE is better in 10 (F2, F7, F8, F14, F16, F17, F19-F22), 18 cases (F2, F3, F7, F8, F10-F17, F21, F22, F24-F27) and worse in 3 (F9, F23,

F28) and 5 cases (F6, F9, F19, F23, F28) respectively. It is observed that on F16 (Rotated Katsuura Function), F17 (Lunacek Bi_Rastrigin Function) and F22 (Composition Function), EaDE is superior or comparable in all the 10-D, 30-D, 50-D and 100-D cases while there is no function in which EaDE consistently loses to Variant-Sa and Variant-SaM.

To have an in-depth insight into the working processes of the adaptation methods, the percentage of strategies used on 10-D, 30-D, 50-D and 100-D F13, F16 and F22 is shown in Fig. 9.

As seen from Fig. 9:

(1) overall, the three adaptation methods exhibit different patterns with different percentages of strategies;

(2) on the same function, each method shows a similar pattern of percentages of strategies on different dimensionalities;

(3) on different functions, EaDE exhibits different patterns and is more significant than the other two methods. For both Variant-Sa and Variant-SaM, the percentage of SCSS-L-CIPDE_GD0.9 is larger than SCSS-L-SHADE_GD0.1 on all the three considered functions. However, for EaDE, it is not the case on F16.

Table 9 Comparison results of different adaptation methods on 10-D, 30-D, 50-D and 100-D CEC2013 benchmark set

|  | EaDE vs. | Variant-Sa | Variant-SaM |
|---|---|---|---|
| 10-D | win | 5 | 6 |
|  | tie | 20 | 20 |
|  | lose | 3 | 2 |
| 30-D | win | 6 | 16 |
|  | tie | 21 | 11 |
|  | lose | 1 | 1 |
| 50-D | win | 10 | 18 |
|  | tie | 15 | 5 |
|  | lose | 3 | 5 |
| 100-D | win | 10 | 16 |
|  | tie | 12 | 9 |
|  | lose | 6 | 3 |

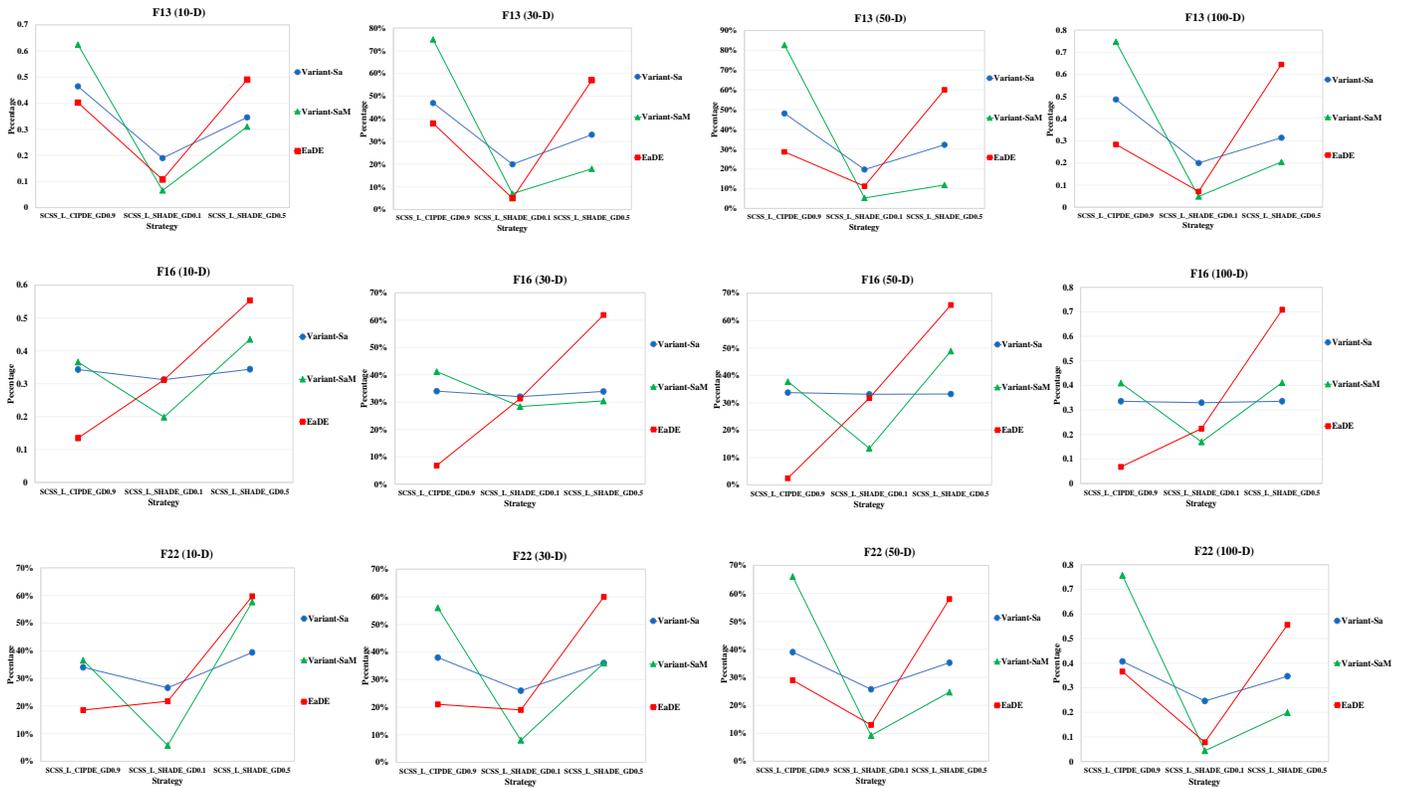

Fig. 9 Percentage of the strategies used in each method on 10-D, 30-D, 50-D and 100-D F13, F16 and F22.

## 4.4 Performance of EaDE: Comparison with state-of-the-art evolutionary algorithms (EAs) and swarm intelligence-based algorithms (SIs)

### 4.4.1 Comparison with state-of-the-art DE algorithms

To demonstrate the performance of EaDE, eight state-of-the-art DE algorithms are considered as baselines, namely SaDE [9], CoDE [15], MPEDE [11], CIPDE [31], jDE [34], JADE [35], L-SHADE [30] and jSO [36]. Tables S4-S7 present the detailed results for 10-D, 30-D, 50-D and 100-D cases, respectively. From the results summarized in Table 10, the followings can be observed:

(1) EaDE performs better than three multi-strategy DEs, i.e. SaDE, CoDE and MPEDE. For instance, in 30-D case, the "win/lose" number is "25/1", "20/1" and "22/1".

(2) EaDE performs better than the rest five single-strategy DEs on the majority of functions. Similarly, take 30-D case as an example, the "win/lose" metric compared with CIPDE, jDE, JADE, L-SHADE and jSO is "17/2", "21/2", "20/1", "17/0" and "15/3" respectively. As is known, L-SHADE and jSO are state-of-the-art. jSO improves the performance of L-SHADE by fine-tuning the mutation factor $F$, crossover factor $CR$ and population size $NP$. However, it is seen that without tuning these parameters, EaDE could still outperform them. The improvement process from L-SHADE to EaDE is L-SHADE → SCSS-L-SHADE_GD0.5→EaDE. The observation that EaDE loses on few cases when compared with
20

L-SHADE indicates that techniques in improvement process could appropriately deal with different kinds of functions in most cases.

(3) With respect to the function types, Table 10 shows that EaDE is consistently more suitable for solving unimodal, basic multimodal and composition functions than the other competitors. With respect to the function dimensionalities, EaDE performs better in all the cases.

Table 11 reports the functions on which the DEs reached the target error value 1.00E-08 in the 30-D case. It is found that EaDE achieves the maximum number of 11 functions. On F12 and F13, only EaDE reached 1.00E-08.

Tables S8-11 and 12 also collect the comparison results on the CEC2014 [37] test suite. Similarly, it is clearly observed that EaDE performs better, winning in 125 (=15+13+24+15+20+19+8+11), 164 (=27+20+20+21+23+24+17+12), 179 (=28+23+25+23+24+25+19+12) and 165 (=27+22+23+22+21+23+17+10) cases and losing in 11 (=1+5+0+2+1+0+0+2), 13 (=0+2+1+3+1+1+0+5), 22 (=0+5+1+3+2+2+2+7) and 32 (=2+4+5+5+5+4+0+7) cases in 10-D, 30-D, 50-D and 100-D respectively. Considering the function types, it is seen from Table 12 that EaDE is superior for solving unimodal, basic multimodal and composition functions while jSO is more suitable for the hybrid functions.

Table 10 Comparison results of EaDE with state-of-the art EAs on 10-D, 30-D, 50-D and 100-D CEC2013 benchmark set

| | win/tie/lose | SaDE | CoDE | MPEDE | CIPDE | jDE | JADE | L-SHADE | jSO |
|---|---|---|---|---|---|---|---|---|---|
| 10-D | Unimodal Functions | 3/2/0 | 1/4/0 | 1/4/0 | 0/5/0 | 2/3/0 | 1/4/0 | 0/5/0 | 0/4/1 |
| | Basic Multimodal Functions | 13/2/0 | 9/4/2 | 14/1/0 | 8/7/0 | 12/2/1 | 12/3/0 | 8/6/1 | 7/6/2 |
| | Composition Functions | 4/3/1 | 5/2/1 | 5/2/1 | 1/6/1 | 4/3/1 | 4/3/1 | 2/5/1 | 1/5/2 |
| | **Total** | **20/7/1** | **15/10/3** | **20/7/1** | **9/18/1** | **18/8/2** | **17/10/1** | **10/16/2** | **8/15/5** |
| 30-D | Unimodal Functions | 3/2/0 | 3/2/0 | 3/2/0 | 3/2/0 | 3/2/0 | 3/2/0 | 0/5/0 | 0/4/1 |
| | Basic Multimodal Functions | 14/0/1 | 11/3/1 | 12/2/1 | 8/5/2 | 12/1/2 | 11/3/1 | 12/3/0 | 9/4/2 |
| | Composition Functions | 8/0/0 | 6/2/0 | 7/1/0 | 6/2/0 | 6/2/0 | 6/2/0 | 5/3/0 | 6/2/0 |
| | **Total** | **25/2/1** | **20/7/1** | **22/5/1** | **17/9/2** | **21/5/2** | **20/7/1** | **17/11/0** | **15/10/3** |
| 50-D | Unimodal Functions | 3/2/0 | 3/2/0 | 3/2/0 | 3/2/0 | 3/2/0 | 3/2/0 | 2/3/0 | 2/2/1 |
| | Basic Multimodal Functions | 14/0/1 | 13/1/1 | 13/1/1 | 11/1/3 | 12/1/2 | 12/1/2 | 13/2/0 | 10/4/1 |
| | Composition Functions | 8/0/0 | 6/1/1 | 7/1/0 | 8/0/0 | 6/2/0 | 7/0/1 | 6/2/0 | 3/3/2 |
| | **Total** | **25/2/1** | **22/4/2** | **23/4/1** | **22/3/3** | **21/5/2** | **22/3/3** | **21/7/0** | **15/9/4** |
| 100-D | Unimodal Functions | 3/2/0 | 3/2/0 | 3/2/0 | 3/2/0 | 3/2/0 | 3/2/0 | 1/3/1 | 2/2/1 |
| | Basic Multimodal Functions | 13/0/2 | 11/2/2 | 8/3/4 | 8/2/5 | 9/2/4 | 10/1/4 | 9/6/0 | 10/3/2 |
| | Composition Functions | 7/1/0 | 8/0/0 | 6/1/1 | 6/1/1 | 8/0/0 | 6/1/1 | 5/3/0 | 4/0/4 |
| | **Total** | **23/3/2** | **22/4/2** | **17/6/5** | **17/5/6** | **20/4/4** | **19/4/5** | **15/12/1** | **16/5/7** |

Table 11 Functions on which an algorithm reached the target error value 1.00E-08 in at least one run of 51 trials in the 30-D case

| | SaDE | CoDE | MPEDE | CIPDE | jDE | JADE | L-SHADE | jSO | EaDE |
|---|---|---|---|---|---|---|---|---|---|
| Fun No. | F1, F5, F11 | F1, F5, F10, F11 | F1-F6, F10, F11 | F1, F3, F5, F6, F11 | F1, F5, F10, F11, F14 | F1, F3, F5, F6, F11, F14 | F1-F6, F10, F11, F14 | F1-F6, F10, F11 | F1-F6, F10-F14 |
| Total | 3 | 4 | 8 | 5 | 5 | 6 | 9 | 8 | **11** |

Table 12 Comparison results of EaDE with state-of-the art EAs on 10-D, 30-D, 50-D and 100-D CEC2014 benchmark set

| | win/tie/lose | SaDE | CoDE | MPEDE | CIPDE | jDE | JADE | L-SHADE | jSO |
|---|---|---|---|---|---|---|---|---|---|
| 10-D | Unimodal Functions | 0/3/0 | 0/3/0 | 0/3/0 | 0/3/0 | 0/3/0 | 0/3/0 | 0/3/0 | 0/3/0 |
| | Basic Multimodal Functions | 9/3/1 | 9/3/1 | 12/1/0 | 6/6/1 | 9/3/1 | 10/3/0 | 6/7/0 | 6/6/1 |
| | Hybrid Functions | 2/4/0 | 0/3/3 | 6/0/0 | 3/2/1 | 5/1/0 | 4/2/0 | 0/6/0 | 2/3/1 |
| | Composition Functions | 4/4/0 | 4/3/1 | 6/2/0 | 6/2/0 | 6/2/0 | 5/3/0 | 2/6/0 | 3/5/0 |
| | **Total** | **15/14/1** | **13/12/5** | **24/6/0** | **15/13/2** | **20/9/1** | **19/11/0** | **8/22/0** | **11/17/2** |
| 30-D | Unimodal Functions | 1/2/0 | 1/2/0 | 0/3/0 | 2/1/0 | 1/2/0 | 2/1/0 | 0/3/0 | 0/3/0 |
| | Basic Multimodal Functions | 13/0/0 | 9/3/1 | 10/3/0 | 7/4/2 | 10/3/0 | 10/3/0 | 8/5/0 | 9/4/0 |
| | Hybrid Functions | 6/0/0 | 5/1/0 | 6/0/0 | 6/0/0 | 6/0/0 | 6/0/0 | 5/1/0 | 1/2/3 |
| | Composition Functions | 7/1/0 | 5/2/1 | 4/3/1 | 6/1/1 | 6/1/1 | 6/1/1 | 4/4/0 | 2/4/2 |
| | **Total** | **27/3/0** | **20/8/2** | **20/9/1** | **21/6/3** | **23/6/1** | **24/5/1** | **17/13/0** | **12/13/5** |
| 50-D | Unimodal Functions | 3/0/0 | 3/0/0 | 2/1/0 | 2/1/0 | 2/1/0 | 2/1/0 | 1/2/0 | 1/2/0 |
| | Basic Multimodal Functions | 12/1/0 | 9/2/2 | 11/2/0 | 9/1/3 | 10/2/1 | 10/1/2 | 10/3/0 | 10/3/0 |
| | Hybrid Functions | 6/0/0 | 5/0/1 | 5/0/1 | 6/0/0 | 6/0/0 | 6/0/0 | 4/1/1 | 0/2/4 |
| | Composition Functions | 7/1/0 | 6/0/2 | 7/1/0 | 6/2/0 | 6/1/1 | 7/1/0 | 4/3/1 | 1/4/3 |
| | **Total** | **28/2/0** | **23/2/5** | **25/4/1** | **23/4/3** | **24/4/2** | **25/3/2** | **19/9/2** | **12/11/7** |
| 100-D | Unimodal Functions | 3/0/0 | 3/0/0 | 2/1/0 | 2/1/0 | 3/0/0 | 2/1/0 | 1/2/0 | 0/3/0 |
| | Basic Multimodal Functions | 11/1/1 | 7/3/3 | 8/1/4 | 6/2/5 | 8/2/3 | 8/1/4 | 8/5/0 | 8/4/1 |
| | Hybrid Functions | 5/0/1 | 6/0/0 | 6/0/0 | 6/0/0 | 5/1/0 | 6/0/0 | 4/2/0 | 1/2/3 |
| | Composition Functions | 8/0/0 | 6/1/1 | 7/0/1 | 8/0/0 | 5/1/2 | 7/1/0 | 4/4/0 | 1/4/3 |
| | **Total** | **27/1/2** | **22/4/4** | **23/2/5** | **22/3/5** | **21/4/5** | **23/3/4** | **17/13/0** | **10/13/7** |

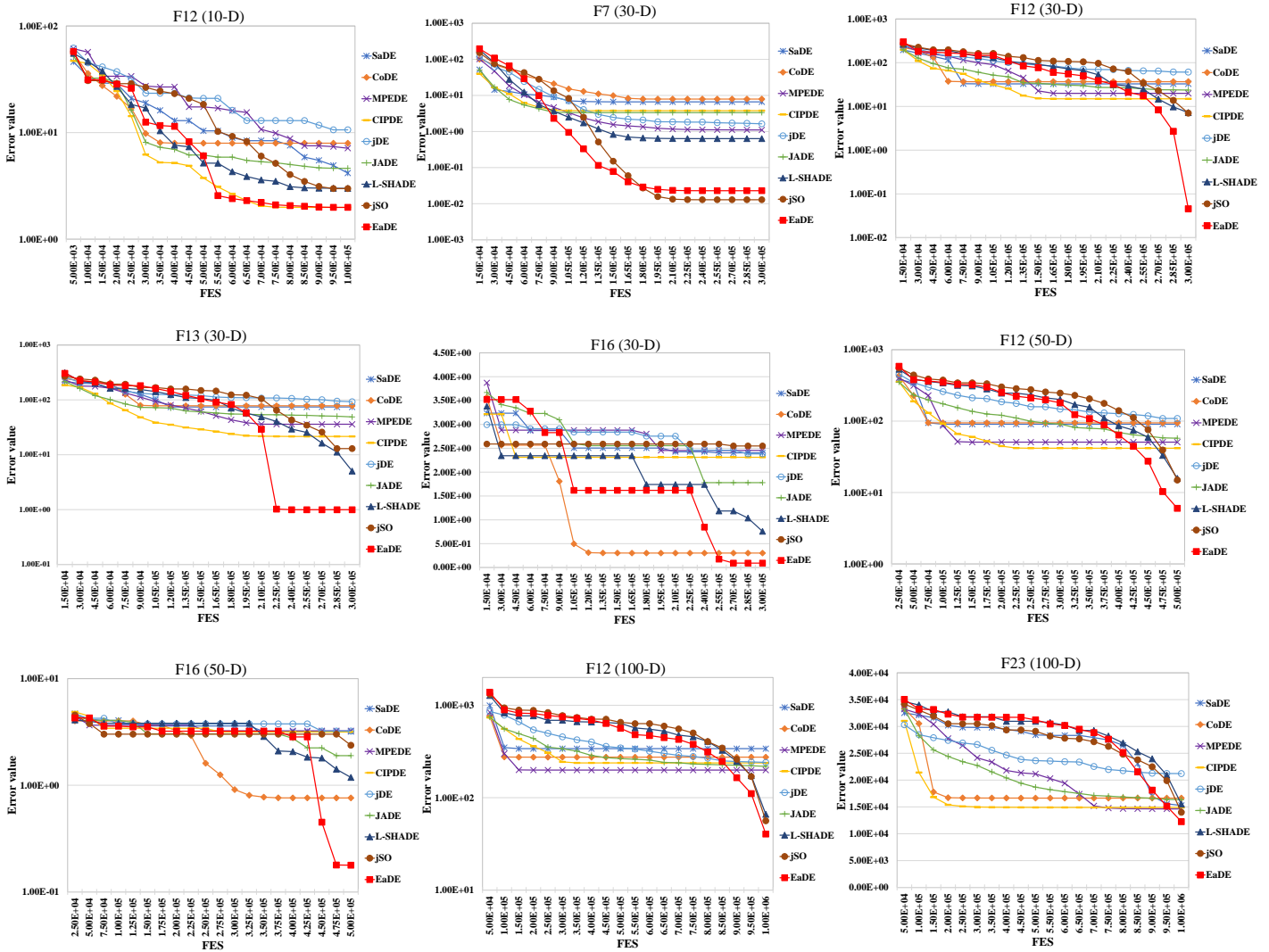

Fig. 10 Convergence plots of the considered DE algorithms on selected CEC2013 functions in the median run.

The convergence graphs of the considered DEs on selected CEC 2013 and CEC 2014 functions are shown in Figs. 10 and S1 respectively. For the CEC2013 functions, jSO achieved the best result on 30-D F7, followed by EaDE. On the rest eight functions, EaDE found the best solutions. It is seen that L-SHADE, jSO and EaDE with the most competitive performance generally converge slower than most of the rest DEs at the early stages. The reason is that at these stages, their population sizes are relatively large. Nevertheless, it could help maintain the population diversity, which is critical for solving the complicated multi-modal functions. Similar observations can also be found on the CEC2014 functions from Fig. S1.

### 4.4.2 Comparison with other state-of-the-art evolutionary algorithms and swarm intelligence-based algorithms

EaDE is further compared with four other algorithms, including the continuous Non-revisiting Genetic Algorithm (cNrGA) [38], the Dynamic Multi-swarm Differential Learning Particle Swarm Optimizer (DMSDL-PSO) [39] and two CMA-ES [40] variants, namely Increasing Population size-based CMA-ES (IPOP-CMA-ES) [41] and Hybrid Sampling Evolution Strategy (HS-ES) [42].

The results on CEC2013 and CEC2014 benchmark suites are presented in Tables S12 and S13 and summarized in Tables 13 and 14, respectively. From these tables:

(1) EaDE performs better than cNrGA and DMSDL-PSO on all considered dimensionalities. For example, in the 30-D case, the "win/lose" metric is "26/0" and "21/2" on CEC2013 and "29/1" and "22/1" on CEC2014;

(2) The two CMA-ES variants have unique advantages in solving some functions, e.g. F9, F15, F23, F24 and F27 from CEC2013 due to their fast convergence. On most of the rest CEC2013 functions, they perform worse than or similar to EaDE. Considering the dimensionality, EaDE outperforms IPOP-CMA-ES on all the cases. While compared with HS-ES, EaDE has advantages in solving the 10-D, 30-D and 50-D CEC2013 functions. In the 100-D case, HS-ES performs slightly better. Similar observations can also be found from the results on CEC2014 functions. Overall, EaDE is competitive against these two state-of-the-art CMA-ESs.

Table 13 Comparison results of EaDE with other EAs and SIs on 10-D, 30-D, 50-D and 100-D CEC2013 benchmark set

| EaDE vs. | | cNrGA | DMSDL-PSO | IPOP-CMA-ES | HS-ES |
|---|---|---|---|---|---|
| 10-D | win | 28 | 19 | 16 | 13 |
| | tie | 0 | 5 | 8 | 12 |
| | lose | 0 | 4 | 4 | 3 |
| 30-D | win | 26 | 21 | 12 | 15 |
| | tie | 2 | 5 | 9 | 6 |
| | lose | 0 | 2 | 7 | 7 |
| 50-D | win | 26 | 21 | 14 | 16 |
| | tie | 1 | 4 | 6 | 3 |
| | lose | 1 | 3 | 8 | 9 |
| 100-D | win | 19 | 16 | 14 | 11 |
| | tie | 7 | 5 | 6 | 3 |
| | lose | 2 | 7 | 8 | 14 |

Table 14 Comparison results of EaDE with other EAs and SIs
on 10-D, 30-D, 50-D and 100-D CEC2014 benchmark set

| EaDE vs. | | cNrGA | DMSDL-PSO | IPOP-CMA-ES | HS-ES |
|---|---|---|---|---|---|
| 10-D | win | 30 | 25 | 24 | 16 |
| | tie | 0 | 3 | 4 | 6 |
| | lose | 0 | 2 | 2 | 8 |
| 30-D | win | 29 | 22 | 23 | 16 |
| | tie | 0 | 7 | 6 | 7 |
| | lose | 1 | 1 | 1 | 7 |
| 50-D | win | 27 | 21 | 23 | 10 |
| | tie | 2 | 4 | 4 | 7 |
| | lose | 1 | 5 | 3 | 13 |
| 100-D | win | 26 | 18 | 22 | 12 |
| | tie | 1 | 2 | 4 | 2 |
| | lose | 3 | 10 | 4 | 16 |

## 4.5 Comparison on real-world problems

The performance of EaDE is further demonstrated by comparison with five state-of-the-art competitors, including cNrGA, DMSDL-PSO, L-SHADE, jSO and HS-ES on eight CEC2011 [43] real-world problems (RWP). Descriptions of the problems are given in Table 15. Thirty trials were performed for each problem with each trial assigned $10000 \times D$ function evaluations. Note that our PC could not afford the memory requirement of cNrGA for solving RWP5 and its result for this problem is not available.

From Table 16, it is clear that EaDE exhibits the best performance among the compared algorithms. The "win/lose" metric is "7/0", "5/1", "5/0", "5/1" and "8/0" when compared with cNrGA, DMSDL-PSO, L-SHADE, jSO and HS-ES, respectively. The advantage is consistent with what have been observed previously on the CEC2013 and CEC2014 benchmarks. Meanwhile, considering that these problems are with a wide range of dimensionalities, the scalability of EaDE is confirmed.

Table 15 Descriptions of the considered CEC2011 problems

| No. | Problem | D |
|---|---|---|
| RWP1 | Parameter Estimation for Frequency Modulated (FM) Sound Waves | 6 |
| RWP2 | Lennard-Jones Potential Problem | 30 |
| RWP3 | Tersoff Potential for model Si (B) | 30 |
| RWP4 | Tersoff Potential for model Si (C) | 30 |
| RWP5 | DED instance 2 | 216 |
| RWP6 | Hydrothermal Scheduling instance 1 | 96 |
| RWP7 | Messenger: Spacecraft Trajectory Optimization Problem | 26 |
| RWP8 | Cassini 2: Spacecraft Trajectory Optimization Problem | 22 |

Table 16 Performance comparisons of EaDE with other competitors
on the eight CEC2011 real-world problems

|  | cNrGA | DMSDL-PSO | L-SHADE | jSO | HS-ES | EaDE |
|---|---|---|---|---|---|---|
| RWP1 | 11.98 −<br>(8.08) | 6.17 −<br>(5.60) | 0.72 −<br>(2.75) | 0.68 +<br>(2.58) | 21.27 −<br>(3.80) | **0.34**<br>(1.86) |
| RWP2 | -25.21 −<br>(3.18) | -27.50 =<br>(0.66) | -27.79 =<br>(0.50) | -27.57 −<br>(0.57) | -27.11 −<br>(0.83) | **-27.79**<br>(0.45) |
| RWP3 | -32.93 −<br>(1.64) | -36.48 =<br>0.55) | -36.77 −<br>(0.29) | -36.75 =<br>(0.27) | -34.28 −<br>(2.05) | **-36.85**<br>(0.02) |
| RWP4 | -25.13 −<br>(3.54) | **-29.17** +<br>(2.69E-13) | -29.17 −<br>(1.82E-4) | -29.17 −<br>(4.21E-4) | -19.99 −<br>(2.98) | -29.11<br>(0.32) |
| RWP5 | N/A | 1078413.53 −<br>(1473.55) | 1050862.18 −<br>(1207.73) | 1050013.2 −<br>(1101.40) | 1067856.93 −<br>(1604.02) | **1047849.62**<br>(954.07) |
| RWP6 | 981716.22 −<br>(129016.35) | 949039.94 −<br>(2434.53) | 926075.85 −<br>(459.42) | 925908.01 −<br>(511.74) | 939616.22 −<br>(3665.08) | **925564.61**<br>(664.47) |
| RWP7 | 18.35 −<br>(2.60) | 16.68 −<br>(1.73) | 15.53 =<br>(0.66) | **15.34** =<br>(0.79) | 20.33 −<br>(0.93) | 15.58<br>(0.75) |
| RWP8 | 21.91 −<br>(3.66) | 17.93 −<br>(1.96) | 14.35 =<br>(2.10) | 15.99 −<br>(2.71) | 21.88 −<br>(2.47) | **13.35**<br>(2.70) |
| win | 7 | 5 | 5 | 5 | 8 | |
| tie | 0 | 2 | 3 | 2 | 0 | |
| lose | 0 | 1 | 0 | 1 | 0 | |

N/A: not available

### 4.6 More discussions on EaDE

#### 4.6.1 Performance sensitivity to *LEN* and *K*

To investigate the performance sensitivity to parameters *LEN* and *K*, 25 combinations with *LEN* = {10, 30, 50, 70, 90} and *K* = {1, 2, 3, 4, 5} are considered. The overall performance ranking of each combination on 30-D functions by Friedman's test [33] is plotted in Fig. 11. It can be observed that the performance of large *LEN* and *K* values, e.g. 90 and 5 is worse than other combinations. For other settings, it is not very sensitive. Table 17 further shows the single-problem performance comparison with the standard setting, i.e. [30, 2]. As seen again, for large [*LEN*, *K*] values, such as [70, 4], [70, 5], [90, 4] and [90, 5], standard setting tends to outperform them. The reason is that large [*LEN*, *K*] value reduces the number of SCSS and adaptive generations and as a result, the algorithm could not timely adjust to appropriate searching directions.

#### 4.6.2 Number of strategies

In Section 3.1, the guideline for studying the exploitation and exploration capabilities (EEC) of strategies was given. In Section 3.2, three constructed strategies, SCSS-L-SHADE_GD0.5, SCSS-L-SHADE_GD0.1 and SCSS-L-CIPDE_GD0.9 with different amounts of EEC were organized following the Ea scheme. Although only three strategies were considered in the current study, the proposed method can also be extended for $N$ ($N > 3$) strategies. Specifically, in SCSS generations, the entire population can be divided into $N-1$ subpopulations according to fitness with each subpopulation mapped to one strategy. Nevertheless, when there are too many strategies, the system may become complex and some inefficient strategies may waste computational resources [5]. Further investigations would be considered as future works.

Table 17 Comparison results of standard EaDE with other parameter settings [*LEN*, *K*] on 30-D CEC2013 benchmark set

| Standard vs. | [10, 1] | [10, 2] | [10, 3] | [10, 4] | [10, 5] | [30, 1] | [30, 2] | [30, 3] | [30, 4] | [30, 5] |
|---|---|---|---|---|---|---|---|---|---|---|
| win | 2 | 3 | 1 | 0 | 0 | 0 | | 0 | 1 | 0 |
| tie | 25 | 25 | 26 | 28 | 28 | 28 | | 28 | 26 | 28 |
| lose | 1 | 0 | 1 | 0 | 0 | 0 | | 0 | 1 | 0 |
| Standard vs. | [50, 1] | [50, 2] | [50, 3] | [50, 4] | [50, 5] | [70, 1] | [70, 2] | [70, 3] | [70, 4] | [70, 5] |
| win | 2 | 1 | 1 | 2 | 2 | 0 | 2 | 1 | 3 | 4 |
| tie | 26 | 27 | 27 | 26 | 25 | 28 | 26 | 27 | 25 | 24 |
| lose | 0 | 0 | 0 | 0 | 1 | 0 | 0 | 0 | 0 | 0 |
| Standard vs. | [90, 1] | [90, 2] | [90, 3] | [90, 4] | [90, 5] | | | | | |
| win | 3 | 3 | 5 | 6 | 6 | | | | | |
| tie | 25 | 25 | 22 | 22 | 22 | | | | | |
| lose | 0 | 0 | 1 | 0 | 0 | | | | | |

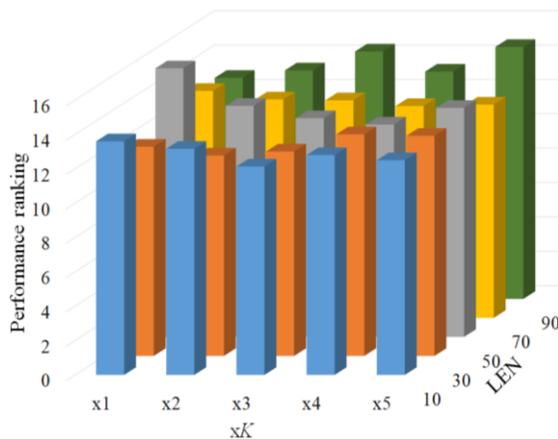

Fig. 11 Performance ranking of different combinations of *LEN* and *K*. The smaller ranking value, the better.

## 5 Conclusion

In this paper, a new strategy adaptation method with explicit exploitation and exploration controls, i.e. Ea scheme has been proposed. Based on the Ea scheme, a new DE named EaDE has been constructed. In EaDE, the evolution process is divided into several SCSS generations and adaptive generations. In SCSS generations, a balanced strategy is employed to detect the exploitation and exploration needs. In adaptive generations, an exploitative strategy or an explorative strategy is adaptively used to meet the needs. To demonstrate the contribution of the Ea scheme, we have compared it with three variants, three components as well as two other popular adaptation methods. We have also further shown the performance of EaDE by comparison with state-of-the art evolutionary algorithms (EAs) and swarm intelligence-based algorithms (SIs) on CEC2013 and CEC2014 test suites as well as on eight real-world problems. Experimental results show that: (1) Ea scheme significantly outperforms the variants, the components and other adaptation methods; and (2) EaDE performs better than state-of-the-art optimization algorithms.

Although further improvements have been achieved with the Ea scheme, we have also noticed that it could not outperform on all the problems. Recognizing that seeking for a best configuration from $3^L$ combinations (with 3 strategies and $L$ generations) is extremely expensive and impractical, we believe the proposed method is a reasonable alternative.

For further study, the proposed Ea scheme might be generalized for other strategies. A possible direction will be to investigate the possibility to extend to other EAs [44] and SIs [45] as well as its application in the hybrids of different EAs and SIs with different exploitation and exploration features.

The MATLAB code of EaDE is available at https://zsxhomepage.github.io/.

**Acknowledgments**

The work was supported in part by City University of Hong Kong under a SRG Grant (Project no: 7004710) and in part by the National Natural Science Foundation of China (No. 61671485).

**References**

[1] M. Črepinšek, S.-H. Liu, M. Mernik, Exploration and exploitation in evolutionary algorithms: a survey, ACM Comput. Surv. 45 (3) (2013) 1–33 Art. No. 35.

[2] M. G. Epitropakis, D. K. Tasoulis, N. G. Pavlidis, V. P. Plagianakos, M. N. Vrahatis, Enhancing differential evolution utilizing proximity-based mutation operators, IEEE Trans. Evol. Comput. 15 (2011) 99–119.

[3] M. Yang, C. Li, Z. Cai and J. Guan, Differential evolution with auto-enhanced population diversity, IEEE Trans. Cybernet. 45 (2015) 302-315.

[4] J. Chacón Castillo, C. Segura, Differential evolution with enhanced diversity maintenance. Optim Lett (2019). https://doi.org/10.1007/s11590-019-01454-5.

[5] G. Wu, R. Mallipeddi, P. N. Suganthan, Ensemble strategies for population-based optimization algorithms – A survey, Swarm Evol. Comput. 44 (2019) 695–711.

[6] R. Storn and K. Price, Differential evolution–A simple and efficient adaptive scheme for global optimization over continuous spaces, Berkeley, CA, Tech. Rep., 1995, tech. Rep. TR-95-012.

[7] S. Das, S. M. Sankha, P.N. Suganthan, Recent advances in differential evolution – An updated survey, Swarm Evol. Comput. 27 (2016) 1-30.

[8] R. D. Al-Dabbagh, F. Neri, N. Idris, M. S. Baba, Algorithm design issues in adaptive differential evolution: review and taxonomy, Swarm Evol. Comput. 43 (2018) 284-311.

[9] A. K. Qin, V. L. Huang, and P. N. Suganthan, Differential evolution algorithm with strategy adaptation for global numerical optimization, IEEE Trans. Evol. Comput. (2009) 398-417.

[10] W. Gong, Z. Cai, C. X. Ling, H. Li, Enhanced differential evolution with adaptive strategies for numerical optimization, IEEE Trans. Systems, Man, and Cybernetics, Part B (Cybernetics) 41 (2010) 397-413.

[11] G. Wu, R. Mallipeddi, P. N. Suganthan, R. Wang, H. Chen, Differential evolution with multi-population based ensemble of mutation strategies, Inf. Sci. 329 (2016) 329–345.

[12] K. Li, Á. Fialho, S. Kwong, Q. Zhang, Adaptive operator selection with bandits for a multiobjective evolutionary algorithm based on decomposition, IEEE Trans. Evol. Comput. 18 (2013) 114-130.

[13] K. Li, Á. Fialho, S. Kwong, Multi-objective differential evolution with adaptive control of parameters and operators, in: C.C. Coello (Ed.), Learning and Intelligent Optimization, 6683 (2011) Springer Berlin Heidelberg, 473–487.

[14] R. Mallipeddi, P. N. Suganthan, Ensemble of constraint handling techniques, IEEE Trans. Evol. Comput. 14 (2010) 561–579.

[15] Y. Wang, Z. Cai, and Q. Zhang, Differential evolution with composite trial vector generation strategies and control parameters, IEEE Trans. Evol. Comput. 15 (2011) 55-66.

[16] W. Gong, A. Zhou, Z. Cai, A multi-operator search strategy based on cheap surrogate models for evolutionary optimization, IEEE Trans. Evol. Comput. 19 (2015) 746–758.

[17] L. Cui, G. Li, Q. Lin, J. Chen, and N. Lu, Adaptive differential evolution algorithm with novel mutation strategies in multiple sub-populations, Comput. Oper. Res. 67 (2016) 155-173.

[18] X. G. Zhou, G. J. Zhang, Differential evolution with underestimation-based multimutation strategy, IEEE Trans. Cybernet. 49 (2019) 1353-1364.

[19] S. X. Zhang, W. S. Chan, Z. K. Peng, S. Y. Zheng, K. S. Tang, Selective-candidate framework with similarity selection rule for evolutionary optimization, Swarm Evol. Comput. 56 (2020) 100696.

[20] S. Das, A. Abraham, U. K. Chakraborty and A. Konar, Differential evolution using a neighborhood-based mutation operator, IEEE Trans. Evol. Comput. 13 (2009) 526-553.

[21] W. Gong, Á. Fialho, Z. Cai, H. Li, Adaptive strategy selection in differential evolution for numerical optimization: an empirical study, Inf. Sci. 181 (2011) 5364-5386.

[22] R. Mallipeddi, P. N. Suganthan, Q. Pan, and M. Tasgetiren, Differential evolution algorithm with ensemble of parameters and mutation strategies, Appl. Soft Comput. 11 (2011) 1679-1696.

[23] Q. Fan, X. Yan, Self-adaptive differential evolution algorithm with zoning evolution of control parameters and adaptive mutation strategies, IEEE Trans. Cybern. 46 (2016) 219-232.

[24] X. Zhou, G. Zhang, Abstract convex underestimation assisted multistage differential evolution, IEEE Trans. Cybernet. 47 (2017) 2730-2741.

[25] S. X. Zhang, S. Y. Zheng, L. M. Zheng, An efficient multiple variants coordination framework for differential evolution, IEEE Trans. Cybernet. 47 (2017) 2780-2793.

[26] S. Gao, Y. Yu, Y. Wang, J. Wang, J. Cheng, and M. Zhou, Chaotic local search-based differential evolution algorithms for optimization, IEEE Trans Systems, Man, and Cybernetics: Systems, (2019), DOI: 10.1109/TSMC.2019.2956121

[27] G. Sun, Y. Cai, T. Wang, H. Tian, C. Wang, Y. Chen, Differential evolution with individual-dependent topology adaptation, Inf. Sci. 450 (2018) 1-38.

[28] M. Tian, X. Gao. Differential evolution with neighborhood-based adaptive evolution mechanism for numerical optimization, Inf. Sci. 478 (2019) 422-448.

[29] S. X. Zhang, L. M. Zheng, K. S. Tang, S. Y. Zheng, and W. S. Chan, Multi-layer competitive-cooperative framework for performance enhancement of differential evolution, Inf. Sci. 482 (2019) 86-104.

[30] R. Tanabe, A.S. Fukunaga, Improving the search performance of shade using linear population size reduction, in: Evolutionary Computation (CEC), 2014 IEEE Congress on, IEEE, 2014, pp. 1658–1665.

[31] L. M. Zheng, S. X. Zhang, K. S. Tang, S. Y. Zheng, Differential evolution powered by collective information, Inf. Sci. 399 (2017) 13–29.

[32] J. J. Liang, B. Y. Qu, P. N. Suganthan, and A. G. Hernández-Díaz, Problem definitions and evaluation criteria for the CEC2013 special session on real-parameter optimization, Comput. Intell. Lab., Zhengzhou Univ., Zhengzhou, China, and Nanyang Technol. Univ., Singapore, Tech. Rep. 201212, Jan. 2013.

[33] J. Derrac, S. García, D. Molina and F. Herrera, A practical tutorial on the use of nonparametric statistical tests as a methodology for comparing evolutionary and swarm intelligence algorithms, Swarm Evol. Comput. 1 (2011) 3–18.

[34] J. Brest, S. Greiner, B. Boskovic, M. Mernik, V. Zumer, Self-adapting control parameters in differential evolution: a comparative study on numerical benchmark problems, IEEE Trans. Evol. Comput. 10 (2006) 646–657.

[35] J. Zhang, A. C. Sanderson, JADE: adaptive differential evolution with optional external archive, IEEE

Trans. Evol. Comput. 13 (2009) 945–958.

[36] J. Brest, M. S Maučec, B. Bošković, Single objective real-parameter optimization: algorithm jSO, in: Proc of the IEEE Congress on Evolutionary Computation, San Sebastian, 2017, pp. 1311–1318.

[37] J. J. Liang, B. Y. Qu , P. N. Suganthan, Problem definitions and evaluation criteria for the CEC 2014 special session and competition on single objective real-parameter numerical optimization, Computational Intelligence Laboratory, Zhengzhou University, Zhengzhou China and Technical Report, Nanyang Technological University, Singapore (2013).

[38] Y. Lou, S. Y. Yuen, Non-revisiting genetic algorithm with adaptive mutation using constant memory, Memetic Comput. 8 (2016) 189–210.

[39] Y. Chen, L. Li, H. Peng, J. Xiao, and Q. Wu, Dynamic multi-swarm differential learning particle swarm optimizer, Swarm Evol. Comput. 39 (2018) 209–221.

[40] N. Hansen, A. Ostermeier, Completely derandomized self-adaptation in evolution strategies, Evol. Comput. 9 (2001) 159–195.

[41] A. Auger, N. Hansen, A restart CMA evolution strategy with increasing population size, in: Proc of the IEEE Congress on Evolutionary Computation, Sep. 2005, pp. 1769–1776.

[42] G. Zhang, Y. Shi, Hybrid sampling evolution strategy for solving single objective bound constrained problems, in Proc. IEEE Congr. Evol. Comput., Rio de Janeiro, 2018, DOI: 10.1109/CEC.2018.8477908.

[43] S. Das, P.N. Suganthan, Problem definitions and evaluation criteria for CEC 2011 competition on testing evolutionary algorithms on real world optimization problems, Jadavpur University, Nanyang Technological University, Technical Report, 2010.

[44] J. Del Ser, E. Osaba, D. Molina, X.S. Yang, S. Salcedo-Sanz, D. Camacho, S. Das, P. N. Suganthan, C.A.C. Coello, and F. Herrera, Bio-inspired computation: Where we stand and what's next, Swarm Evol. Comput. 48 (2019) 220–250.

[45] A. Altan, S. Karasu, Recognition of COVID-19 disease from X-ray images by hybrid model consisting of 2D curvelet transform, chaotic salp swarm algorithm and deep learning technique. Chaos, Solitons & Fractals, 140 (2020) 110071.